\documentclass[10pt,twocolumn,letterpaper]{article}

\usepackage{cvpr}              

\def\cvprPaperID{*****} 
\def\confName{CVPR}
\def\confYear{2023}

\usepackage{graphicx}
\usepackage{amsmath}
\usepackage{amssymb}
\usepackage{booktabs}
\usepackage{bbding}

\usepackage[accsupp]{axessibility}

\usepackage{xcolor}
\newcommand{\tb}[1]{{\textbf{#1}}}

\usepackage[pagebackref,breaklinks,colorlinks]{hyperref}
\usepackage{makecell}

\usepackage[capitalize]{cleveref}
\crefname{section}{Sec.}{Secs.}
\Crefname{section}{Section}{Sections}
\Crefname{table}{Table}{Tables}
\crefname{table}{Tab.}{Tabs.}

\def\cvprPaperID{5008} 
\def\confName{CVPR}
\def\confYear{2023}

\begin{document}

\title{\LaTeX\ Author Guidelines for \confName~Proceedings}
\title{Grid-guided Neural Radiance Fields for Large Urban Scenes}

\author{Linning Xu$^{\star 1}$,
	Yuanbo Xiangli$^{\star 1}$,
	Sida Peng$^4$,
	Xingang Pan$^3$, \\
	Nanxuan Zhao$^5$, 
	Christian Theobalt$^3$, 
	Bo Dai$^2$\Envelope, 
	Dahua Lin$^{1,2}$\\
	$^1$The Chinese University of Hong Kong,
	$^2$Shanghai AI Laboratory, \\
	$^3$Max Planck Institute for Informatics, 
	$^4$Zhejiang University,
	$^5$Adobe Research \\
	\tt\small
	\{xl020,xy019,dhlin\}@ie.cuhk.edu.hk,
	\{xpan,theobalt\}@mpi-inf.mpg.de,\\
	\tt\small
	pengsida@zju.edu.cn, nanxuanzhao@gmail.com, daibo@pjlab.org.cn \\
	\\
}

\twocolumn[{%
	\renewcommand
	\twocolumn[1][]{#1}%
	\maketitle
	\begin{center}
		\centering
		\vspace{-30pt}
		\includegraphics[width=\textwidth]{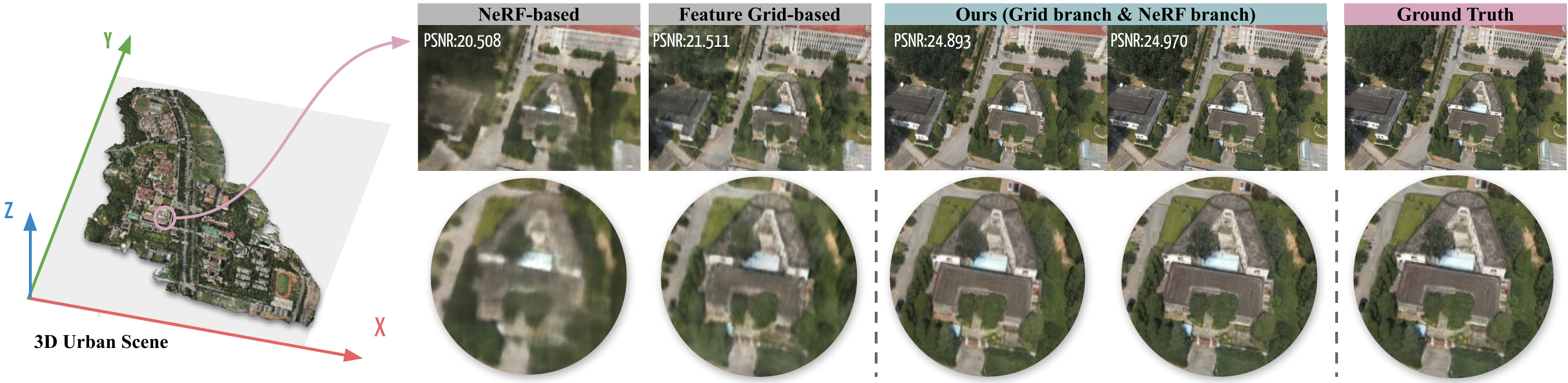}
		\vspace{-15pt}
		\captionof{figure}{ \small We perform large urban scene rendering with a novel grid-guided neural radiance fields. 
		An example of our target large urban scene is shown on the left,
		which spans over 2.7$km^2$ ground areas captured by over $5k$ drone images. 
		We show that the rendering results from NeRF-based methods~\cite{mildenhall2020nerf,turki2021mega} 
		are blurry and overly smoothed with limited model capacity, while feature grid-based methods~\cite{chen2022tensorf,mueller2022instant} 
		tend to display noisy artifacts when adapting to large-scale scenes with high-resolution feature grids. 
		Our proposed two-branch model combines the merits from both approaches and achieves photorealistic novel view renderings with remarkable improvements over existing methods. Both two branches gain significant enhancements over their individual baselines. (Project page: \href{https://city-super.github.io/gridnerf/}{ https://city-super.github.io/gridnerf})}
		\label{fig:teaser}
	\end{center}
}]


\begin{abstract}
\label{sec:abs}
Purely MLP-based neural radiance fields (NeRF-based methods) often suffer from underfitting with blurred renderings on large-scale scenes due to limited model capacity.
Recent approaches propose to geographically divide the scene and adopt multiple sub-NeRFs to model each region individually, leading to linear scale-up in training costs and the number of sub-NeRFs as the scene expands.
An alternative solution is to use a feature grid representation,
which is computationally efficient and can naturally scale to a large scene with increased grid resolutions.
However, the feature grid tends to be less constrained and often reaches suboptimal solutions, producing noisy artifacts in renderings, especially in regions with complex geometry and texture. 
In this work, we present a new framework that 
realizes high-fidelity rendering on large urban scenes while being computationally efficient.
We propose to use a compact multi-resolution ground feature plane representation to coarsely capture the scene, 
and complement it with positional encoding inputs through another NeRF branch for rendering in a joint learning fashion. 
We show that such an integration can utilize the advantages of two alternative solutions: a light-weighted NeRF is sufficient, under the guidance of the feature grid representation, 
to render photorealistic novel views with fine details; 
and the jointly optimized ground feature planes, can meanwhile gain further refinements, forming a more accurate and compact feature space and output much more natural rendering results.
\end{abstract}

\vspace{-10pt}

\section{Introduction}
\label{sec:intro}

Large urban scene modeling has been drawing lots of research attention with the recent emergence of neural radiance fields (NeRF) due to its photorealistic rendering and model compactness~\cite{tewari2021advances,mildenhall2020nerf,barron2021mipnerf360,tancik2022block,xiangli2022bungeenerf,turki2021mega}. Such modeling can enable a variety of practical applications,  including autonomous vehicle simulation~\cite{Li2019AADSAA,Yang2020SurfelGANSR,Ost2021NeuralSG}, aerial surveying~\cite{Du2018TheUA,Bozcan2020AUAIRAM}, and embodied AI~\cite{Morad2021EmbodiedVN,Truong2021BiDirectionalDA}.
NeRF-based methods have shown impressive results on object-level scenes with their \emph{continuity prior} benefited from the MLP architecture and \emph{high-frequency details} with the globally shared positional encodings. However, they often fail to model large and complex scenes. These methods suffer
 from underfitting due to limited model capacity and only produce blurred renderings without fine details~\cite{xiangli2022bungeenerf,tancik2022block,turki2021mega}.
BlockNeRF~\cite{tancik2020fourfeat} and MegaNeRF~\cite{turki2021mega} propose to geographically divide urban scenes and assign each region a different sub-NeRF to learn in parallel.
Subsequently,
when the scale and complexity of the target scene increases,
they inevitably suffer from a trade-off between the number of sub-NeRFs and the capacity required by each sub-NeRF to fully capture all the fine details in each region.
Another stream of grid-based representations represents the target scene using a grid of features~\cite{liu2020nsvf,yu2021plenoctrees,yu2021plenoxels,mueller2022instant,liu2020nsvf,Martel2021ACORNAC,Takikawa2021NeuralGL}. 
These methods are generally much faster during rendering and more efficient when the scene scales up.
However, as each cell of the feature grid is individually optimized in a locally encoded manner, the resulting feature grids tend to be less continuous across the scene compared to NeRF-based methods.
Although more flexibility is intuitively beneficial for capturing fine details, the lack of inherent continuity makes this representation vulnerable to suboptimal solutions with noisy artifacts, as demonstrated in Fig.~\ref{fig:teaser}.

To effectively reconstruct large urban scenes with implicit neural representations, in this work, we propose a two-branch model architecture that takes a unified scene representation that integrates both \textbf{grid-based} and \textbf{NeRF-based} approaches under a joint learning scheme.
Our key insight is that these two types of representations 
can be used complementary to each other:
while feature grids can easily fit local scene content with explicit and independently learned features,
NeRF introduces an inherent \emph{global continuity} on the learned scene content with its shareable MLP weights across all 3D coordinate inputs.
NeRF can also encourage capturing high-frequency scene details by matching the \emph{positional encodings} as Fourier features with the bandwidth of details.
However, unlike feature grid representation, NeRF is less effective in compacting large scene contents into its globally shared latent coordinate space.

Concretely,
we firstly model the target scene with a feature grid in a pre-train stage, which coarsely captures scene geometry and appearance. The coarse feature grid is then used to
1) guide NeRF's point sampling to let it concentrate around the scene surface;
and 2) supply NeRF's positional encodings with extra features about the scene geometry and appearance at sampled positions.
Under such guidance, NeRF can effectively and efficiently pick up finer details in a drastically compressed sampling space. Moreover, as coarse-level geometry and appearance information are explicitly provided to NeRF, a light-weight MLP is sufficient to learn a mapping from global coordinates to volume densities and color values.
The coarse feature grids get further optimized with gradients from the NeRF branch in the second joint-learning stage,
which regularizes them to produce more accurate and natural rendering results when applied in isolation.
To further reduce memory footprint and learn a reliable feature grid for large urban scenes,
we adopt a compact factorization of the 3D feature grid to approximate it without losing representation capacity.
Based on the observation that essential semantics such as the urban layouts are mainly distributed on the ground (\ie, $xy$-plane), we propose to factorize the 3D feature grid into 2D ground feature planes spanning the scene and a vertically shared feature vector along the $z$-axis.
The benefits are manifold:
1) The memory is reduced from $\mathcal{O}(N^3)$ to $\mathcal{O}(N^2)$.
2) The learned feature grid is enforced to be disentangled into
highly compact ground feature plans, offering explicit and informative scene layouts.
Extensive experiments show the effectiveness of our unified model and scene representation. 
When rendering novel views in practice, users are allowed to use either the grid branch at a faster rendering speed, or the NeRF branch, with more high-frequency details and spatial smoothness, yet at the cost of a relatively slower rendering.
\section{Related Works and Background}
\label{sec:related}

\noindent \textbf{Large-scale Scene Reconstruction and Rendering.}  This is a long-standing problem in computer vision and graphics, and many early works \cite{fruh2004automated, snavely2006phototourism, pollefeys2008detailed, li2008modeling, agarwal2011building,zhu2018very, schonberger2016structure} have attempted to solve it.
Most of these methods adopt a three-stage pipeline. They firstly detect salient points \cite{lowe2004sift, hassner2012sifts} in 2D images and construct the point descriptors \cite{rublee2011orb, bay2006surf}. The point descriptors are then matched across images to obtain 2D correspondences that are used for camera pose estimation and 3D point triangulation. Finally, the camera poses and 3D points are jointly optimized to minimize the difference between the projected and image points using bundle adjustment \cite{hartleyziss2004, triggs1999bundle}.
These methods have shown impressive reconstruction performance on large-scale scenes.
\cite{shan2013turing, losasso2004geometry} utilize the reconstruction results to accomplish the free-viewpoint navigation of scenes.
However, there are often artifacts or holes in the reconstructed scenes, which limit the rendering quality.
Recent methods \cite{li2020crowdsampling, meshry2019neural, martin2021nerf} exploit deep learning techniques to improve the results of image synthesis.
\cite{meshry2019neural} rasterizes the recovered point clouds into deep buffers, which are then interpreted into 2D images using the 2D convolutional neural networks.
More recently, \cite{martin2021nerf} leverages the technique of generative latent optimization to reconstruct radiance fields from photo collections, enabling it to achieve photorealistic rendering results. \cite{devries2021unconstrained,sharma2022seeing} adopted a similar ground plane representation as us for scene generation and static-dynamic disentanglement tasks respectively, yet the scale of their applicable scenes and the rendering qualities remain limited.

\smallskip
\noindent \textbf{Volumetric Scene Representations.}
Coordinate-based multi-layer perceptrons have become a popular representation for 3D shape modeling \cite{park2019deepsdf, chabra2020deep, mescheder2019occupancy, chen2019learning} and novel view synthesis \cite{niemeyer2020differentiable, mildenhall2020nerf, barron2021mip, barron2021mipnerf360}.
To represent high-resolution 3D shapes, some methods adopt MLP networks that take continuous 3D coordinates as input and predict the target values~\cite{park2019deepsdf,mescheder2019occupancy,chibane2020neural}.
\cite{chibane2020implicit, peng2020convolutional, chen2021multiresolution} introduces convolutional neural networks to learn powerful scene priors, enabling the coordinate-based representations to handle larger scenes.
In the field of novel view synthesis, NeRF \cite{mildenhall2020nerf} represents 3D scenes as density and color fields and optimizes this representation from images through volume rendering techniques.
\cite{yu2021plenoctrees, hedman2021baking, liu2020nsvf, reiser2021kilonerf} extend NeRF with efficient data structures to accelerate the rendering process.
\cite{chan2021efficient, schwarz2020graf, niemeyer2021giraffe, xu20213d} combine NeRF with generative models and achieve 3D-aware image generation.
\cite{li2021neural, park2021nerfies, pumarola2021d} augment NeRF with motion fields, enabling them to handle dynamic scenes.

\smallskip
\noindent \textbf{NeRF Scale-up.}
While the aforementioned NeRF approaches mainly consider scenes of a limited scale, scaling up NeRF to handle large-scale scenes such as cities would enable broader applications.
\cite{liu2020nsvf, xu2022point, turki2021mega} have tried to improve the rendering quality of NeRF on large-scale scenes.
Feature grid methods \cite{liu2020nsvf, li2022neural} map input coordinates to a high dimensional space with a lookup from the predefined table of learnable feature vectors, which augments the approximation ability of neural networks.
PointNeRF \cite{xu2022point} regresses the radiance field from the point cloud and accomplishes the high-quality rendering of indoor scenes.
BungeeNeRF \cite{xiangli2022bungeenerf} designs a multi-scale representation that efficiently models the scene content and improves the rendering quality.
\cite{turki2021mega, tancik2022block} decompose the scene into several spatial regions that are separately represented by NeRF networks.
When processing large scenes, another critical problem is how to reduce the training time.
Several techniques such as image encoder \cite{wang2021ibrnet, zhang2022nerfusion}, auto-decoder \cite{ramon2021h3d, dupont2022data} and meta-learning \cite{tancik2021learned, bergman2021fast} are used to pretrain networks on a dataset to learn the scene prior for improving the optimization process.
\cite{sun2021direct, mueller2022instant, yu2021plenoxels,sun2022improved} further explore the grid representations to accelerate the training speed.
TensoRF \cite{chen2022tensorf} investigate the factorization of 3D scenes, enabling it to compactly represent scenes and achieve the fast training. Instant-NGP~\cite{mueller2022instant} adopts a multi-resolution hash table of feature vectors that enables extremely fast renderings. However, we notice both these methods suffer from noisy feature gridss when applied to large scenes.
\section{Grid-guided Neural Radiance Fields}
\label{sec:methodology}

\begin{figure*}
	\centering
	\includegraphics[width=\linewidth]{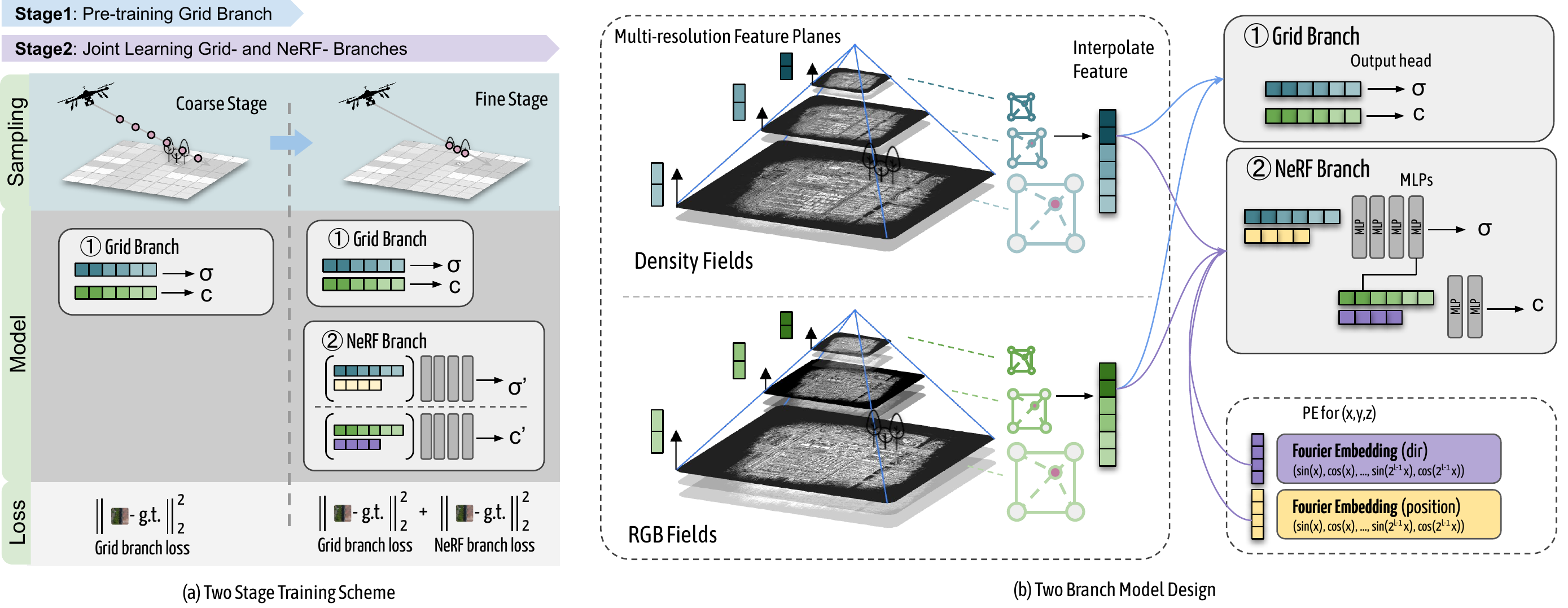}
	\caption{\small \textbf{Overview of our framework}. The core of our model is a novel two-branch structure, namely the grid branch and NeRF branch. 
	1) We start by capturing the scene with a pyramid of feature planes at the pre-train stage, and performing a coarse sampling of ray points and predicting their radiance values through a shallow MLP renderer (grid branch), supervised by the MSE loss on the volumetrically integrated pixel colors. 
	This step yields a set of informative multi-resolution density/appearance feature planes.
	2) Next, we proceed to the joint learning stage and perform a finer sampling. We use the learned feature grid to guide NeRF branch sampling to concentrate on the scene surface. 
	The sampled points' grid feature is inferred by bilinear interpolation on the feature planes. 
	The features are then concatenated with the positional encoding and fed to NeRF branch to predict volume density and color. Note that, the grid branch outputs maintain being supervised with the ground truth images along with the fine-rendering results from the NeRF branch during the joint training.}
	\vspace{-5mm}
	\label{fig:network}
\end{figure*}

Recall that NeRF-based representations obtain the point density and color by passing the positional encoding (PE) of point coordinates into an 8-layer MLP~\cite{mildenhall2020nerf}. Such a model is highly compact as the entire scene contents are encoded in the MLP weights taking PE embedding as inputs, yet they face difficulty in scaling up limited by model capacities. In contrast, grid-based representations encode a scene into a feature grid, which can be intuitively thought of as a 3D voxel grid with grid resolution matched with the actual 3D space. Each voxel stores a feature vector at the vertices and can then be interpolated to extract a feature value at the query point coordinate and converted to the point density and color via a small network. As feature grids are often implemented as high-dimensional tensors, various factorization methods can also be applied to obtain more compact feature grid representations~\cite{chen2022tensorf}.

To effectively represent large urban scenes, we propose the grid-guided neural radiance fields, which combines the expertise of NeRF-based and grid-based methods.  The grid features are enforced to reliably capture as much local information as possible with a \emph{multi-resolution ground feature plane}.
We then let the positional encoded coordinates information pick up the missing high-frequency details and produce high-quality renderings. With a rough construction of feature grid that captrue the scene at multiple resolution,
the density field is also used to guide NeRF's sampling procedure. 
When training the NeRF branch, the grid features are jointly optimized and supervised with the reconstruction loss from both two branches. 

Fig.~\ref{fig:network} illustrates the overall pipeline of our system. 
In Sec.~\ref{subsec:multires_grid_pretrain}, we describe the pre-train of our multi-resolution ground feature plane representation; Sec.~\ref{subsec:grid_guided_nerf} introduces the grid-guided learning of neural radiance fields, corresponding to the NeRF branch in Fig.~\ref{fig:network}; Finally, we elaborate how the NeRF branch helps refine the pre-trained grid feature of the grid branch in Sec.~\ref{subsec:rectified_grid_feature}. 

\subsection{Multi-resolution Feature Grid Pre-train}
\label{subsec:multires_grid_pretrain}
Fig.~\ref{fig:teaser} illustrates a representative scenario of large urban scenes. Inspired by the fact that large urban scenes mainly ground on the $xy$-plane, we propose to represent the target large urban scene with a major \emph{ground feature plane} by constructing a multi-resolution plane-vector feature space. 
Enforcing the ground plane compression gives more informative feature planes compared to the ones obtained from full 3D grids.
This compact representation is especially suitable in urban scene scenarios and behaves robustly to sparse view training data.
Various operations (\eg, concatenation, outer product) can then be considered here to recover the 3D information from the 2D ground feature planes. 
The outer product operation was adopted in~\cite{chen2022tensorf} from the perspective of tensor factorization with low-rank approximation, which achieves a more compressed memory footprint while maintaining high quality.
The volume density $\sigma \in \mathbb{R}^+$ and view-dependent color $c \in \mathbb{R}^3$ grid-planes are separately learned to capture more environmental effects that only influences appearence. 
Formally, our grid-based radiance field is written as:
$\sigma, c = F_{\sigma}(\mathcal{G}_\sigma(X)), F_c(\mathcal{G}_c(X),\operatorname{PE}(d))$
where $\mathcal{G}_\sigma(X) \in \mathbb{R}^{R_\sigma}$, $\mathcal{G}_c(X) \in \mathbb{R}^{R_c}$ are the extracted interpolated feature values from the two grid-planes at location $X\in \mathbb{R}^3$. $F_{\sigma}, F_c$ are two fusing functions, implemented with two small MLP, that translate the concatenated density/appearance features to $\sigma, c$, and
$d \in \mathbb{S}^2$ is the viewing direction.
$\operatorname{PE}$ here represents the positional encoding $(\sin (\cdot), \cos (\cdot), \ldots, \sin \left(2^{L-1} (\cdot)\right), \cos \left(2^{L-1} (\cdot)\right))$ as in \cite{mildenhall2020nerf}.
The grid branch is then trained with $N$ query samples along the ray and predicts the pixel color following the volume rendering process as in~\cite{mildenhall2020nerf}, where the loss is the total squared error between	the rendered and true pixel colors for this coarse sampling stage, as shown in Fig.~\ref{fig:network} (a).

We approximate the full density and appearance grid features with the channel-wise outer product of the ground feature planes $R_{\sigma}$ and $R_c$, as well as the globally encoded $z$-axis feature vectors, following the practice of~\cite{chen2022tensorf}.
For each channel $r\in R_{\sigma}$ and $R_c$, the corresponding tensor grid of features are: where $\mathbf v^{z}$ represents the vector along $z$-axis, $\mathbf{M}^{xy}$ denotes the  matrix spanning $xy$-plane, and $\circ$ represents the outer product.
With the constraints of learning a shared $z$-axis feature vector, the optimized ground feature plane is encouraged to encode sufficient local scene contents, that can be translated by a globally shared MLP renderer.
For a specific grid resolution $n$, the density and the appearance tensors $\mathcal{G}_\sigma^n, \mathcal{G}_c^n$ are then obtained as the concatenation of $R_{\sigma}$, $R_c$ feature components:
	\begin{equation}
		\mathcal{G}_{\sigma}^n = \oplus[ (\mathbf{v}_{\sigma, r}^{z} \circ \mathbf{M}_{\sigma,r}^{xy})]_{R_\sigma}, \quad
		\mathcal{G}_{c}^n = \oplus[(\mathbf{v}_{c, r}^{z} \circ \mathbf{M}_{c,r}^{xy})]_{R_c},
	\end{equation}
where $\oplus$ denotes the concatenation operation over the $R_\sigma$ and $R_c$ dimension.
To capture different degrees of scene local complexity, 
we learn a multi-resolution feature grid with $\mathcal{G}_\sigma = \{\mathcal{G}_\sigma^n\}$ and $\mathcal{G}_c = \{\mathcal{G}_c^n\}$.
The yielding multi-resolution feature grid contains features at different granularity to describe the scene, which is particularly suitable for urban environments with objects appearing in different scales. 

\subsection{Grid-guided Neural Radiance Field}
\label{subsec:grid_guided_nerf}
A NeRF trained from scratch is required to reason about the whole scene from purely positional inputs, which only provides a band of Fourier frequencies in PE. For large urban scenes that naturally bear a wide range of granularity for geometry and texture details, NeRF constantly biases towards learning low-frequency functions, as pointed out in~\cite{tancik2020fourfeat,xiangli2022bungeenerf}. This problem gets amplified in large scenes where a large amount of information needs to be encoded. To remedy this, we propose to compress the sampling space of NeRF with the pre-trained feature grid density and enrich NeRF's pure coordinates inputs with the coarse grid features initialized in the pre-train stage.
	
Despite being of limited accuracy and granularity, the pre-trained grid feature can already offer an approximation of the scene which can be used to 1) guide NeRF's point sampling and 2) provide intermediate features as a complement to the coordinate inputs. As demonstrated in Fig.~\ref{fig:network}, instead of mapping coordinates spanning the entire sample space, NeRF can now concentrate on the approximated scene surface for more efficient and denser point sampling, and evoke high-frequency Fourier features in positional encoding to recover finer details.
Meanwhile, points along the sampled ray are projected onto the multi-resolution feature planes to retrieve density and appearance features via bilinear interpolation. The inferred grid features are then concatenated to the positional encoding as input to NeRF branch. The per-point density and color $\sigma^{\prime}, c^{\prime}$ are predicted via the NeRF branch network $F^\prime$ as:
\begin{equation}
	(\sigma^{\prime}, c^{\prime}) = F^{{\prime}}(\mathcal{G}_\sigma(X), \mathcal{G}_c(X),\operatorname{PE}(X), \operatorname{PE}(d)).
	\label{eqn:rf}
 \end{equation}
The multi-resolution feature plane plays a critical role as it provides information about the scene at multiple granularities, relieving the fitting burden of NeRF's PE so that it can concentrate on refining the fine details of the scene.
Particularly, while a high grid resolution can guarantee that each voxel in space to captures its local contents, the quality grows at the cost of storage regardless of the possible heterogeneity of detail level across the scene. It is therefore more efficient to provide such high-frequency details with Fourier features that only cost several dimensions in PE and can be adapted to scene throughout the learning process.
	
Note that the two-branch supervision and the two-stage training are necessary as: (1) a randomly initialized feature grid can hardly provide informative scene contents and may entangle the role of two types of network inputs. (2) The pre-train stage is much faster than the one with NeRF branch included, making it more efficient to reliably construct a coarse geometry with grid branch only. (3) Unlike~\cite{sun2021direct} which freezes the voxel grids when supplying PE inputs, we will later show that the feature grids can gain further refinement with its jointing learning with the NeRF branch. Moreover, as the grid branch is also supervised with reconstruction loss, it enforces the grid branch to continue enriching its captured scene information where the PE input can focus on the missing high-frequency details.

\subsection{Refined Grid Feature Planes from NeRF}
\label{subsec:rectified_grid_feature}
\begin{figure*}[t!]
	\centering
	\includegraphics[width=\linewidth]{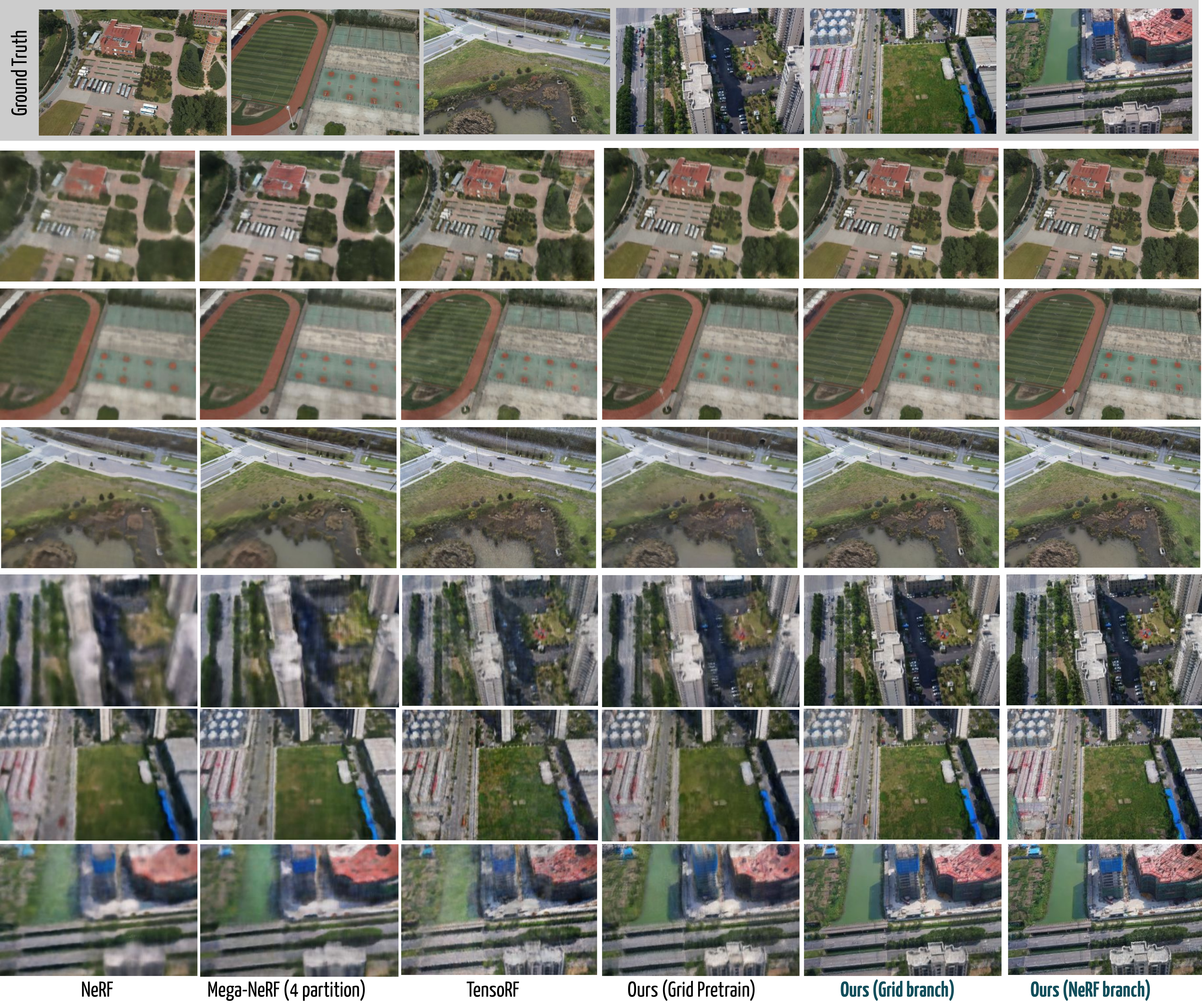}
	\vspace{-20pt}
	\caption{\small Qualitative comparison between baselines and ours. 
	On large urban scenes, MLP-based methods (NeRF) and (Mega-NeRF) suffer from severe blurry artifacts. 
	Grid-based method (TensoRF) shows better results, but tend to produce noisy appearance with inaccurate shapes. 
	The pre-trained multi-resolution grid feature (Ours, Grid Pretrain) improves over the single high-resolution methods, yet the results is still suboptimal.
	Our final model achieves photo-realistic quality compared with ground-truth images on novel views. While grid branch (Ours, Grid branch) and NeRF branch (Ours, NeRF branch) receive similar metric scores, it is preferred to render from the NeRF branch which has sharper details and smoother spatial continuity, especially when rendering long videos in practice.
}
    \vspace{-15pt}
	\label{fig:comparison}
\end{figure*}

Recall that the feature grid relies on bilinear interpolation on the ground feature plane to obtain a feature vector of points within a voxel. The mechanism can yield detailed reconstruction results given sufficiently high grid resolution, such that the finest variation in the scene can be recovered. However, learning a grid with matching resolution can be highly memory-consuming for large urban scenes, as indicated in~\cite{mueller2022instant}. Moreover, the grid feature lacks the incentive to capture accurate variations within a voxel with merely reconstruction loss on ground truth RGB.
We therefore jointly optimize the feature plane and vector with NeRF to enhance the supervision signal for grid features with point-wise guidance from the supplied NeRF inputs. Another benefit NeRF brings is the global regularization on the independently optimized grid features. Fig.~\ref{fig:teaser} and Fig.~\ref{fig:comparison} show that grid-based methods suffer from noisy artifacts because of the lack of constraints on space continuity and semantic similarity. NeRF, on the contrary, uses a shared MLP for the entire scene space. We will later show that the rendered novel views interpreted from the grid branch can get largely improved after its joint training with the NeRF branch.

\section{Experiments}
\label{sec:experiment}

\subsection{Experimental Setup}

\noindent \textbf{Dataset.} Our main experiments are conducted on the real-world urban scenes. The three scenes cover different urban environments including rural rubble site~~\cite{turki2021mega} (\emph{Rubble}), university campus (\emph{Campus}), and residential complexes (\emph{Residential}). 
The camera poses are obtained from the photogrammetry software \href{https://www.bentley.com/en/terms-of-use-and-select-online-agreement}{ContextCapture}.  
Additional experiements and results on general scene datasets, alterative camera poses, and techniques for further improvements can be found in the supplementary and our webpage.

\smallskip
\noindent \textbf{Baselines and Implementations.}
We compare the performance of our approach with 1) NeRF~\cite{mildenhall2020nerf} applied on the whole scene; 2) Mega-NeRF~\cite{turki2021mega} with $4$ partitions; 3) TensoRF~\cite{chen2022tensorf}, which reduces the memory footprint of feature grid via low-rank tensor factorization and considered suitable for large-scene scenarios. 
For NeRF and Mega-NeRF, we accordingly adopted larger model with $12$ layers and $256$ hidden units. 
The highest frequency of position encoding is set to $2^{15}$, inserted to NeRF model via skip connection at the $4,6,8,10$ layers. 
We use hierarchical sampling during training with $64$ coarse and $128$ fine samples per ray. 
All NeRF models are optimized using Adam optimizer~\cite{kingma2014adam} with a learning rate that decayed exponentially from $5e^{-4}$ and a batch size of $2048$ rays, trained for $150k$ iterations. 
For TensoRF, in accordance with our observation on large urban scenes discussed previously in Sec.~\ref{subsec:multires_grid_pretrain}, we evaluated the simplified version that factorizes a feature grid into an xy-plane matrix and z-axis vector components. $16/48$ components are used for density and appearance feature grid respectively.
Starting from an initial low-resolution grid with $128^3$ voxels, the grid gets upsampled to $1024^3$ linearly in logarithmic space during training. 
The grid resolution along each dimension is scaled by the $x,y,z$ dimensions. 
A small MLP with $2$ fully connected layers of $128$ hidden layers and ReLU activation is used as the color output head.
Adam optimizer is adopted with initial leaning rate of $0.02$ for tensor factors and $0.01$ for the MLP decoder. The batch size is $4096$. The model is trained for $100k$ iterations.

Our method takes the matching grid resolution as the highest resolution feature plane and $8/16$ components for density/appearance grid respectively, with another two at the downsampled $\times 4$ and $\times 16$ resolution. The MLP head for the grid branch is same as TensoRF. The NeRF branch use a 4 MLP layer without skip layer. The highest frequency of position encoding is also set to $2^{15}$. Adam optimizer is adopted with initial learning rate of $0.02$ for tensor factors and $0.01$ for the MLP layers with batch size $4096$.
We pre-train the grid branch for the first $10k$ iterations and joint optimization for another $100k$ iterations, and the time fraction between two stages is roughly 1:4.
We use weighted loss 1:1 for two branches in joint training. 

\subsection{Results Analysis}
We report the performance of baselines and our method in Fig.~\ref{fig:comparison} and Tab.~\ref{tab:compare} both qualitatively and quantitatively. A significant improvement can be observed in visual quality and across all metrics.
Our method reveals sharper geometry and more delicate details than purely MLP-based approaches (NeRF and Mega-NeRF).
Especially, due to the limited capacity and spectral bias of NeRF, it always fails to model rapid changes in geometry and color, such as vegetation and stripes on the playground. 
Even though geographically partitioning scenes into small regions slightly helps as shown in the Mega-NeRF baseline, the rendered results still appear to be overly smoothed.
On the contrary, with the guidance from the learned feature grid, NeRF's sampling space is effectively and drastically compressed to near scene surface.
The density and appearance features sampled from the ground feature planes explicitly indicate the scene contents, as depicted in Fig.~\ref{fig:comparison}. Despite being less accurate, it already offers informative local geometry and texture, and encourages NeRF's PE to pick up the missing scene details.

\begin{table}[t!]
	\caption{\small Quantitative comparison on three large urban scene datasets. We report PSNR($\uparrow$), LPIPS($\uparrow$)~\cite{zhang2018unreasonable}, SSIM($\downarrow$) metric on the test views. The \tb{best} and \underline{second best} results are highlighted.}
	\vspace{-4mm}
	\label{tab:compare}
	\begin{center}
		\resizebox{\linewidth}{!}{
			\begin{tabular}{l|rrr|rrr|rrr}
				\toprule
				Scene   &   \multicolumn{3}{c|}{\emph{Rubble}}  &   \multicolumn{3}{c|}{\emph{Campus}} &   \multicolumn{3}{c}{\emph{Residential}} \\
				\midrule
				Metric &  PSNR & LPIPS & SSIM   &
				PSNR & LPIPS & SSIM   &
				PSNR & LPIPS & SSIM     \\
				\midrule
				NeRF   & 21.659 & 0.541 & 0.547 & 22.283 & 0.560 & 0.509 & 18.548 & 0.622 & 0.401 \\
				Mega-NeRF & 23.505 & 0.516 & 0.565 & 22.365 & 0.496 & 0.544 & 19.350 & 0.561 & 0.452 \\
				TensoRF   & 23.800 & 0.478 & 0.670 & 20.915 & 0.471 & 0.571 & 18.332 & 0.575 & 0.428 \\
				Ours (pretrain)   & 22.617 & 0.451 & 0.622 & 24.542 & 0.385 & 0.698 & 21.032 & 0.428 & 0.620\\
				\midrule
				\tb{Ours} (grid branch) & \tb{25.467} & \underline{0.213} & \tb{0.780} & \tb{25.505} & \underline{0.174} & \tb{0.767} & \tb{24.372} & \underline{0.142} & \tb{0.807}\\
				\tb{Ours} (nerf branch) & \underline{24.130} & \tb{0.207} & \underline{0.767} & \underline{24.903} &  \tb{0.162}  & \underline{0.757} & \underline{23.765} & \tb{0.137} & \underline{0.802}\\
				\bottomrule
			\end{tabular}		
		}
	\vspace{-3mm}
	\end{center}
	\centering
\end{table}

\begin{figure}[t!]
	\centering
	\includegraphics[width=\linewidth]{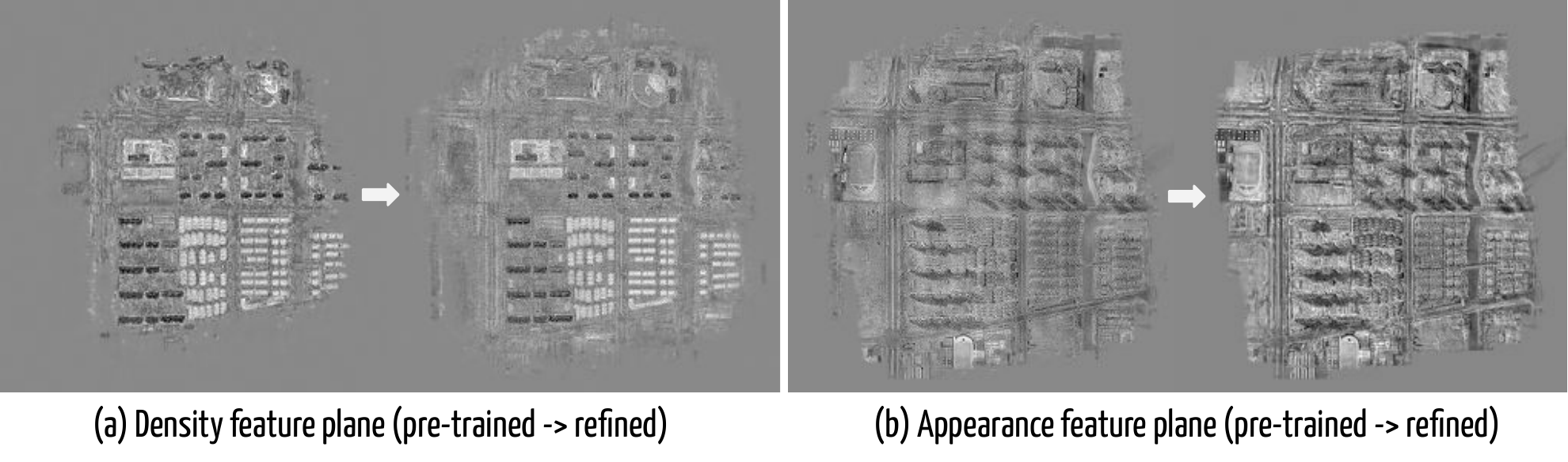}
	\vspace{-6mm}
	\caption{\small Visualization of one feature component in (a) density and (b) appearance feature plane (\emph{Residential} scene). Compared to the pre-trained feature planes, the refined ones are less noisy; sharper edges and regular shapes of grouped objects can also be clearly identified. Since density and appearance features are independently learned, they encode different information that describes the scene. The appearance feature can capture environmental effects like shadows, as shown in (b).}
	\vspace{-5mm}
\label{fig:rectified}
\end{figure}

\begin{figure}[t!]
	\centering
	\includegraphics[width=\linewidth]{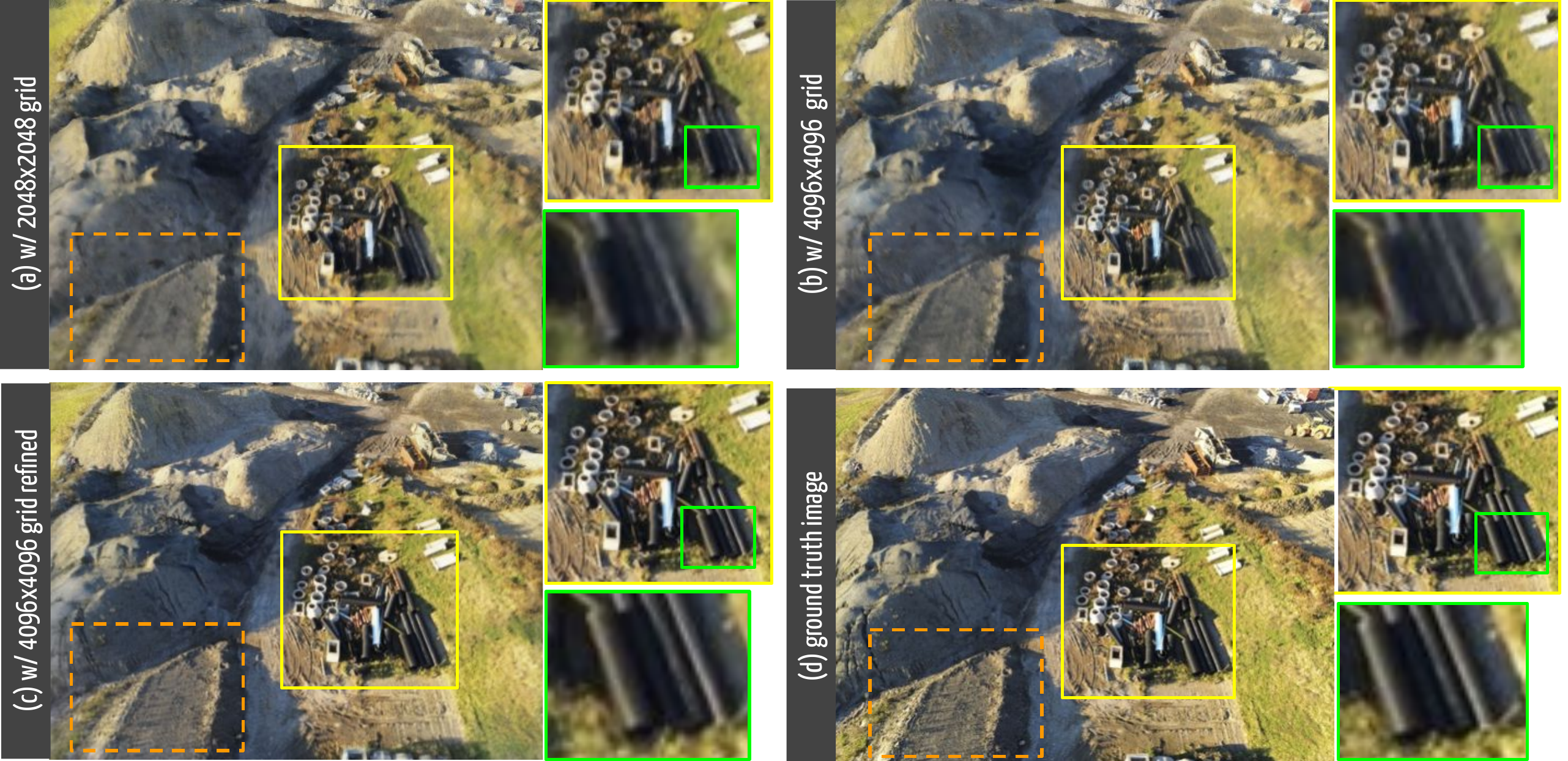}
	\caption{\small Qualitative comparison showing the rendering results using features learned (a) at a moderate grid resolution ($2048^2$), (b) at a high grid resolution ($4096^2$) and (c) from the re grid branch at resolution ($4096^2$). Despite higher grid resolution leads to better visual quality, adding NeRF supervision pushes the quality toward photorealistic one step further.}
	\vspace{-3mm}
	\label{fig:grid-rectified}
\end{figure}

\smallskip
\noindent \textbf{Refined Ground Feature Planes.}
Grid-based methods often require explicitly imposed regularization, such as the total variation loss or L1 loss~\cite{chen2022tensorf}, to avoid noises in regions with fewer observations, otherwise the independently optimized grid features can easily result in fuzzy and wavy appearances, as illustrated in Fig.~\ref{fig:comparison}.
By jointly optimizing with the NeRF branch, the $xy$-plane and $z$-axis encoding are constantly improved to encode more local details while becoming less noisy. A drastic improvement in fidelity can be observed in Fig.~\ref{fig:grid-rectified}. 
Similar refinement can also be observed in feature space. Take one dimensional density plane on the \emph{Residential} scene (Fig.~\ref{fig:rectified}) as an example,
while a coarse floor layout of the targeting urban area can already be identified on the pre-trained $xy$-plane feature (Fig.~\ref{fig:rectified}(a)), it still missed details like sharp edges and varied shapes and colors widespread in the scene, which are hard to be represented with grids unless finer grid resolution is adopted. 
NeRF, on the other hand, searches for scene surfaces with points, which provides a more accurate  and meaningful signal for grid feature optimization, and boosts it out of local minima. 
Another noticeable artifact on the pre-trained feature plane is the noise on continuous regions (\eg, land, facade) due to grid variation, which is largely eased after jointly optimized with NeRF.
This can be ascribed to NeRF's continuous representation of the scene, which imposes an implicit regularization on the feature grid by constructing a stronger correlation among coordinates.
The resulting refined feature planes (Fig.~\ref{fig:rectified}(b)) exhibit smooth grid features with a cleaner silhouette where content-similar grids can be clustered together (\eg, buildings, duplexes, and roads). 

\begin{figure}[t!]
	\centering
	\includegraphics[width=0.9\linewidth]{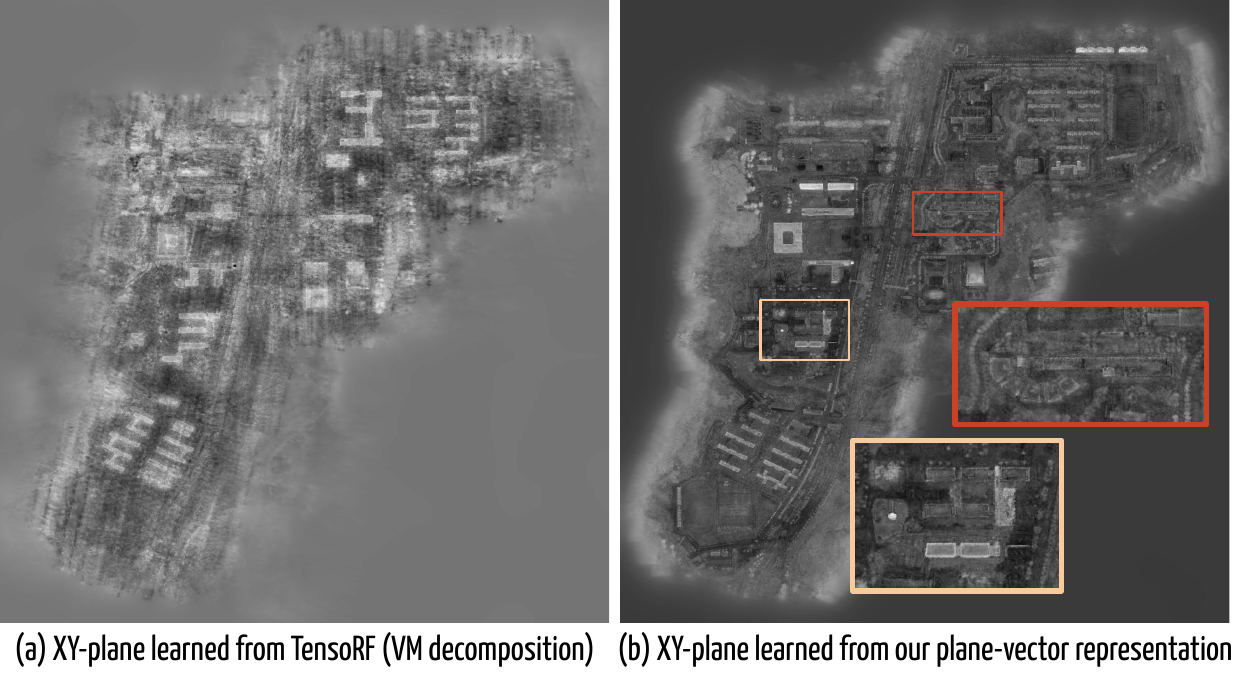}
	\vspace{-3mm}
	\caption{\small Visualization of a slice of $xy$ feature plane from (a) TensoRF's factorization; and (b) our ground plane representation. 
		Our joint learning results in more accurate plane features with sharp region boundaries that is better
		aligned with the scene’s physical ground plan, which is naturally more suitable for large urban scene modeling and downstream analysis. A cleaner feature grid also reveals that the learned latent space is more \emph{compact}, which is critical for large-scale modeling even with limited model capacity.}
	\vspace{-10pt}
	\label{fig:xy-plane}
\end{figure}

\smallskip
\noindent \textbf{Compact Representations. } 
While it is instinctive to design a heavy framework for modeling large scenes, our principle is to keep it compact and efficient without significantly reduced quality. 
Bearing that in mind, we modeled the full 3D feature grid with a succinct plane-vector representation. We demonstrate in Fig.~\ref{fig:xy-plane} that,
with similar performance on reconstructing large urban scenes (PSNR: (TensoRF) 21.075 vs. (Ours) 20.915),
the 2D ground plane learned from TensoRF's VM decomposition~\cite{chen2022tensorf} appears to be fuzzier and less informative than ours; 
and our representation uses less parameters ($3e8$) compared to TensoRF ($4e8$).  
Moreover, recall that grid resolution is critical for purely voxel-based representation to obtain high quality renderings, our method realizes photorealistic rendering of large scenes without further upsampling. Although supplying finer-grained feature planes to our framework is beneficial, the integration with NeRF largely alleviates the dependence on grid resolution to capture scene details.
From NeRF's perspective, we show that a relative small MLP is sufficient to handle large scenes by taking the learned grid features with PE, and achieves superior result than the scaled-up NeRF and Mega-NeRF~\cite{turki2021mega}, as shown in Fig.~\ref{fig:comparison}. 

\subsection{Ablations}
Ablations are conducted to verify the impact of 1) different model configurations: for the grid branch, we switch to single resolution feature grid at different resolutions; for NeRF branch, we inspect model capacity and the frequency bandwidth of PE in helping NeRF recover the scene details; 
Apart from the model architecture, we also look into 2) the efficacy of enriching NeRF's pure coordinated input with grid features; and 3) the efficacy of NeRF as a supervision signal to enhance feature grid.

\begin{figure}[t!]
	\centering
	\includegraphics[width=0.84\linewidth]{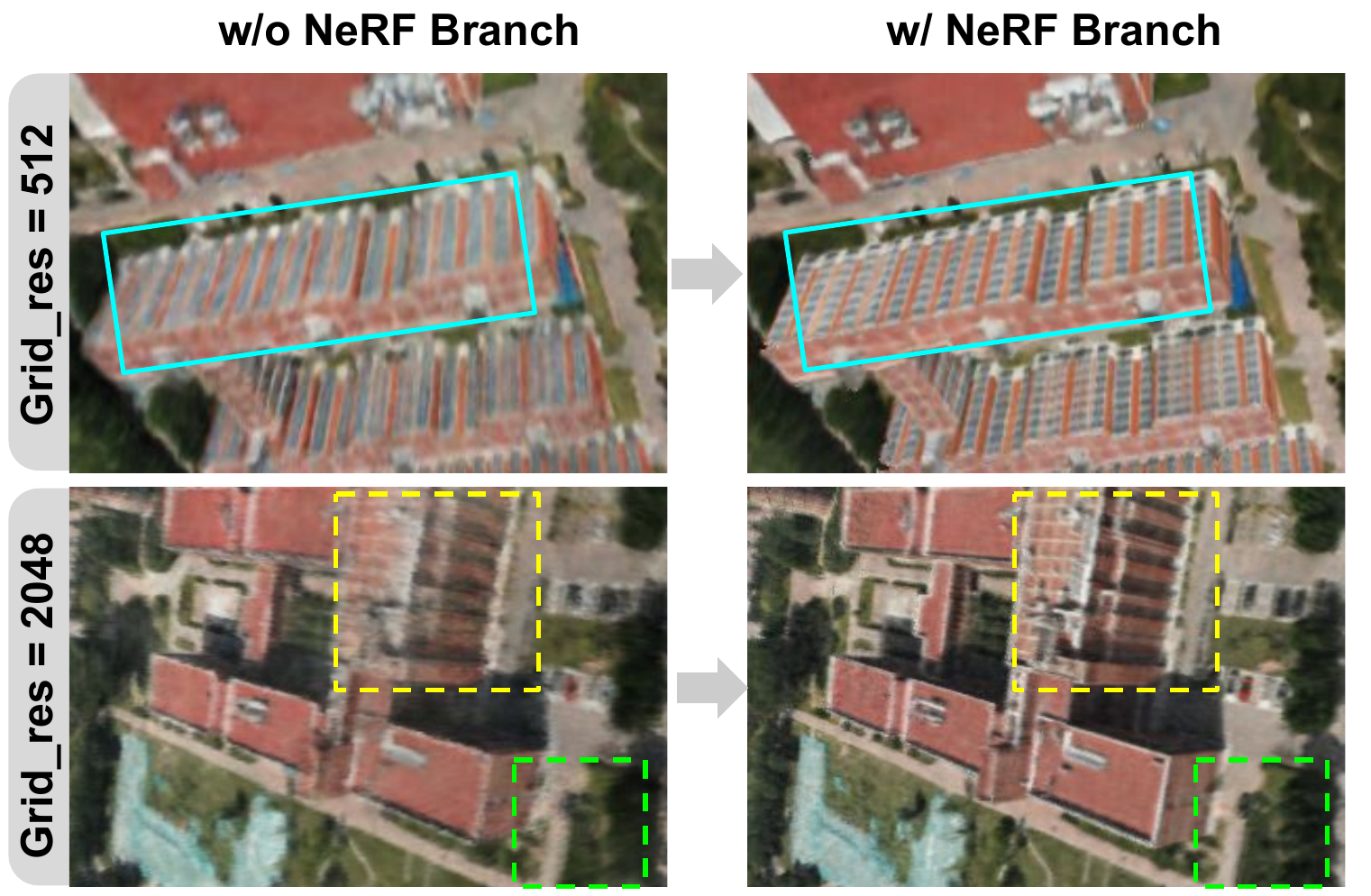}
	\caption{\small Using a single resolution grid feature results in inferior results. Low-resolution feature grid suffers from blurry artifacts and high-resolution grid gives noisy results. NeRF branch greatly helps remedy these issues with its point-wise supervision signal to pick up more details and global prior to regularize grid features. }
	\vspace{-3mm}
	\label{fig:abl_gridres}
\end{figure}

\begin{figure}[t!]
	\centering
	\includegraphics[width=0.85\linewidth]{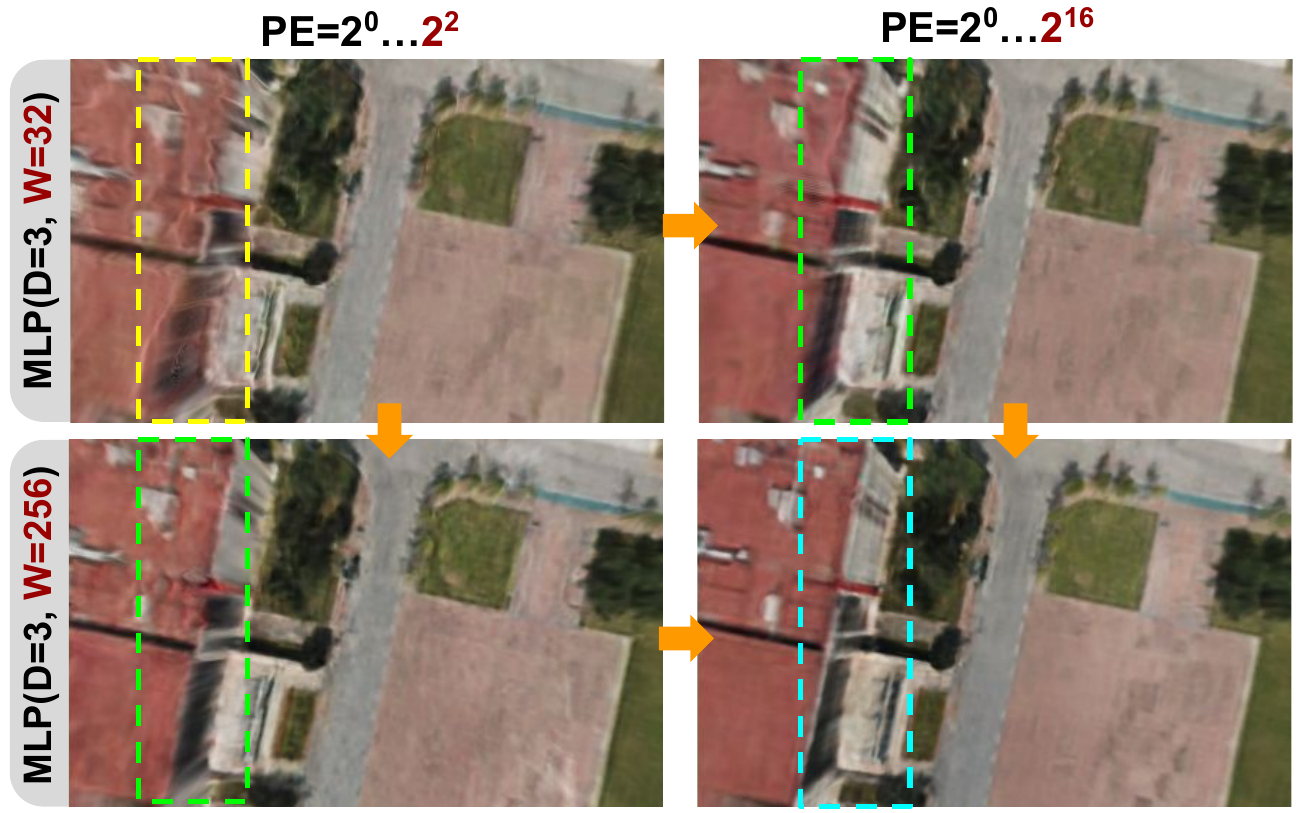}
	\caption{\small A small renderer is insufficient to translate the grid features, producing inaccurate geometry and scene contents. However, given enough modeling capacity, NeRF picks up more details with the help of increasingly higher PE frequency channels.}
	\vspace{-3mm}
	\label{fig:abl_nerfsize}
\end{figure}

\smallskip
\noindent\textbf{Model configuration.}
For the grid branch, we show in Fig.~\ref{fig:abl_gridres} that adopting a single resolution feature grid leads to inferior performance. Concretely, results from a low resolution ($512^2$) grid branch already suffers from blurry artifacts during pre-training. Adding NeRF branch at the latter stage can help produce more details to the facade and rooftop but still lack sharp detail in general. On the other hand, results from a high resolution ($2048^2$) grid branch is fuzzy and noisy at the pre-training stage, which is eased by NeRF to a large extent, but still unstable at continuous regions, such as roads and walls. 
Existing works~\cite{sun2021direct,mueller2022instant,chen2022tensorf} usually adopt a small MLP as a renderer to translate grid features. In Fig.~\ref{fig:abl_nerfsize} We show that for our scenario, NeRF with a small model capacity ($D$=3, $W$=32) is insufficient to translate grid features of such complex scenes, giving inaccurate geometry and missing a large amount of scene details. Naively increasing frequency bandwidth in PE does little help under this circumstance.
By enlarging the MLP ($D$=3, $W$=256), significant improvements can be observed, from which imposing higher frequency inputs via PE can help recover more scene details.

\smallskip
\noindent \textbf{Efficacy of grid features to NeRF. }
We start by simply supplying NeRF with grid features without tuning grid features and supervision from the grid branch. NeRF can already benefit from the local features encoded in the grid features, with $\sim1db$ improvement in PSNR. Tuning the feature grid can further achieve $\sim2.5db$ gain in PSNR.

\smallskip
\noindent \textbf{Efficacy of NeRF supervision to feature grid.}
As depicted in Fig.~\ref{fig:abl_gridres}, NeRF helps feature grids in recovering more details when grid resolution is inadequate, and smooth unregularized features with global prior to produce more consistent rendering results. On high-resolution grid, combining NeRF can raise PSNR by $\sim2db$ on the \emph{Campus} scene. 
\section{Discussion and Conclusion}
\label{sec:conclusion}
In this work, we target on large urban scene rendering, and  propose a novel framework that integrates the MLP-based NeRF with an explicitly constructed a feature grid to effectively encode both local and global scene informations. Our method overcomes various drawbacks of state-of-the-art methods when when applied to large-scale scenes.
Our model achieves high visual fidelity rendering even for extremely large-scale urban scenes, which is crucial for real-world application scenarios.

While we mainly investigate the ground feature plane representation tailored for large urban scene scenario, our two-branch design can also be considered for other grid-based representation, serving as an additional regularization on the learned feature values by bringing more continuity. 
Still, our model inherits some limitations of NeRF-based methods, such as the slow training for our joint-learning stage. Another critical issue is dealing with a large amount of high-resolution images. The current batch sampling of shuffled rays is highly ineffective without distributed training. More discussions can be found in the supplementary. 

\noindent\textbf{Acknowledgment}
This project is funded in part by Shanghai AI Laboratory, CUHK Interdisciplinary AI Research Institute, and the Centre for Perceptual and Interactive Intelligence (CPIl) Ltd under the Innovation and Technology Commission (ITC)'s InnoHK.

{\small
\bibliographystyle{ieee_fullname}
\bibliography{egbib}

\begin{thebibliography}{10}\itemsep=-1pt

\bibitem{agarwal2011building}
Sameer Agarwal, Yasutaka Furukawa, Noah Snavely, Ian Simon, Brian Curless,
  Steven~M Seitz, and Richard Szeliski.
\newblock Building rome in a day.
\newblock {\em Communications of the ACM}, 2011.

\bibitem{barron2021mip}
Jonathan~T Barron, Ben Mildenhall, Matthew Tancik, Peter Hedman, Ricardo
  Martin-Brualla, and Pratul~P Srinivasan.
\newblock Mip-{NeRF}: A multiscale representation for anti-aliasing neural
  radiance fields.
\newblock {\em ICCV}, 2021.

\bibitem{barron2021mipnerf360}
Jonathan~T Barron, Ben Mildenhall, Dor Verbin, Pratul~P Srinivasan, and Peter
  Hedman.
\newblock Mip-nerf 360: Unbounded anti-aliased neural radiance fields.
\newblock {\em arXiv preprint arXiv:2111.12077}, 2021.

\bibitem{bay2006surf}
Herbert Bay, Tinne Tuytelaars, and Luc Van~Gool.
\newblock Surf: Speeded up robust features.
\newblock In {\em ECCV}, 2006.

\bibitem{bergman2021fast}
Alexander Bergman, Petr Kellnhofer, and Gordon Wetzstein.
\newblock Fast training of neural lumigraph representations using meta
  learning.
\newblock {\em Advances in Neural Information Processing Systems}, 34, 2021.

\bibitem{Bozcan2020AUAIRAM}
Ilker Bozcan and Erdal Kayacan.
\newblock Au-air: A multi-modal unmanned aerial vehicle dataset for low
  altitude traffic surveillance.
\newblock {\em 2020 IEEE International Conference on Robotics and Automation
  (ICRA)}, pages 8504--8510, 2020.

\bibitem{chabra2020deep}
Rohan Chabra, Jan~E Lenssen, Eddy Ilg, Tanner Schmidt, Julian Straub, Steven
  Lovegrove, and Richard Newcombe.
\newblock Deep local shapes: Learning local sdf priors for detailed 3d
  reconstruction.
\newblock In {\em European Conference on Computer Vision}, pages 608--625.
  Springer, 2020.

\bibitem{chan2021efficient}
Eric~R Chan, Connor~Z Lin, Matthew~A Chan, Koki Nagano, Boxiao Pan, Shalini
  De~Mello, Orazio Gallo, Leonidas Guibas, Jonathan Tremblay, Sameh Khamis,
  et~al.
\newblock Efficient geometry-aware 3d generative adversarial networks.
\newblock {\em arXiv preprint arXiv:2112.07945}, 2021.

\bibitem{chen2022tensorf}
Anpei Chen, Zexiang Xu, Andreas Geiger, Jingyi Yu, and Hao Su.
\newblock Tensorf: Tensorial radiance fields.
\newblock {\em arXiv preprint arXiv:2203.09517}, 2022.

\bibitem{chen2019learning}
Zhiqin Chen and Hao Zhang.
\newblock Learning implicit fields for generative shape modeling.
\newblock In {\em Proceedings of the IEEE/CVF Conference on Computer Vision and
  Pattern Recognition}, pages 5939--5948, 2019.

\bibitem{chen2021multiresolution}
Zhang Chen, Yinda Zhang, Kyle Genova, Sean Fanello, Sofien Bouaziz, Christian
  H{\"a}ne, Ruofei Du, Cem Keskin, Thomas Funkhouser, and Danhang Tang.
\newblock Multiresolution deep implicit functions for 3d shape representation.
\newblock In {\em Proceedings of the IEEE/CVF International Conference on
  Computer Vision}, pages 13087--13096, 2021.

\bibitem{chibane2020implicit}
Julian Chibane, Thiemo Alldieck, and Gerard Pons-Moll.
\newblock Implicit functions in feature space for 3d shape reconstruction and
  completion.
\newblock In {\em Proceedings of the IEEE/CVF Conference on Computer Vision and
  Pattern Recognition}, pages 6970--6981, 2020.

\bibitem{chibane2020neural}
Julian Chibane, Gerard Pons-Moll, et~al.
\newblock Neural unsigned distance fields for implicit function learning.
\newblock {\em Advances in Neural Information Processing Systems},
  33:21638--21652, 2020.

\bibitem{devries2021unconstrained}
Terrance DeVries, Miguel~Angel Bautista, Nitish Srivastava, Graham~W Taylor,
  and Joshua~M Susskind.
\newblock Unconstrained scene generation with locally conditioned radiance
  fields.
\newblock In {\em Proceedings of the IEEE/CVF International Conference on
  Computer Vision}, pages 14304--14313, 2021.

\bibitem{Du2018TheUA}
Dawei Du, Yuankai Qi, Hongyang Yu, Yi-Fan Yang, Kaiwen Duan, Guorong Li, W.
  Zhang, Qingming Huang, and Qi Tian.
\newblock The unmanned aerial vehicle benchmark: Object detection and tracking.
\newblock In {\em ECCV}, 2018.

\bibitem{dupont2022data}
Emilien Dupont, Hyunjik Kim, SM Eslami, Danilo Rezende, and Dan Rosenbaum.
\newblock From data to functa: Your data point is a function and you should
  treat it like one.
\newblock {\em arXiv preprint arXiv:2201.12204}, 2022.

\bibitem{fruh2004automated}
Christian Fr{\"u}h and Avideh Zakhor.
\newblock An automated method for large-scale, ground-based city model
  acquisition.
\newblock {\em IJCV}, 2004.

\bibitem{hartleyziss2004}
R.~I. Hartley and A. Zisserman.
\newblock {\em Multiple View Geometry in Computer Vision}.
\newblock Cambridge University Press, second edition, 2004.

\bibitem{hassner2012sifts}
Tal Hassner, Viki Mayzels, and Lihi Zelnik-Manor.
\newblock On sifts and their scales.
\newblock In {\em CVPR}, 2012.

\bibitem{hedman2021baking}
Peter Hedman, Pratul~P Srinivasan, Ben Mildenhall, Jonathan~T Barron, and Paul
  Debevec.
\newblock Baking neural radiance fields for real-time view synthesis.
\newblock {\em arXiv:2103.14645}, 2021.

\bibitem{kingma2014adam}
Diederik~P Kingma and Jimmy Ba.
\newblock Adam: A method for stochastic optimization.
\newblock {\em ICLR}, 2015.

\bibitem{li2022neural}
Rui Li, Darius R{\"U}ckert, Yuanhao Wang, Ramzi Idoughi, and Wolfgang Heidrich.
\newblock Neural adaptive scene tracing.
\newblock {\em arXiv preprint arXiv:2202.13664}, 2022.

\bibitem{Li2019AADSAA}
W. Li, C.~W. Pan, R. Zhang, J.~P. Ren, Y.~X. Ma, J. Fang, F.~L. Yan, Q.~C.
  Geng, X.~Y. Huang, H.~J. Gong, W.~W. Xu, G.~P. Wang, Dinesh Manocha, and
  R.~G. Yang.
\newblock Aads: Augmented autonomous driving simulation using data-driven
  algorithms.
\newblock {\em Science Robotics}, 4, 2019.

\bibitem{li2008modeling}
Xiaowei Li, Changchang Wu, Christopher Zach, Svetlana Lazebnik, and Jan-Michael
  Frahm.
\newblock Modeling and recognition of landmark image collections using iconic
  scene graphs.
\newblock {\em ECCV}, 2008.

\bibitem{li2021neural}
Zhengqi Li, Simon Niklaus, Noah Snavely, and Oliver Wang.
\newblock Neural scene flow fields for space-time view synthesis of dynamic
  scenes.
\newblock In {\em Proceedings of the IEEE/CVF Conference on Computer Vision and
  Pattern Recognition}, pages 6498--6508, 2021.

\bibitem{li2020crowdsampling}
Zhengqi Li, Wenqi Xian, Abe Davis, and Noah Snavely.
\newblock Crowdsampling the plenoptic function.
\newblock In {\em European Conference on Computer Vision}, pages 178--196.
  Springer, 2020.

\bibitem{liu2020nsvf}
Lingjie Liu, Jiatao Gu, Kyaw~Zaw Lin, Tat{-}Seng Chua, and Christian Theobalt.
\newblock Neural sparse voxel fields.
\newblock {\em NeurIPS}, 2020.

\bibitem{losasso2004geometry}
Frank Losasso and Hugues Hoppe.
\newblock Geometry clipmaps: terrain rendering using nested regular grids.
\newblock {\em Siggraph}, 2004.

\bibitem{lowe2004sift}
David~G Lowe.
\newblock Distinctive image features from scale-invariant keypoints.
\newblock {\em IJCV}, 2004.

\bibitem{Martel2021ACORNAC}
Julien N.~P. Martel, David~B. Lindell, Connor~Z. Lin, Eric Chan, Marco
  Monteiro, and Gordon Wetzstein.
\newblock Acorn: Adaptive coordinate networks for neural scene representation.
\newblock {\em ACM Trans. Graph.}, 40:58:1--58:13, 2021.

\bibitem{martin2021nerf}
Ricardo Martin-Brualla, Noha Radwan, Mehdi~SM Sajjadi, Jonathan~T Barron,
  Alexey Dosovitskiy, and Daniel Duckworth.
\newblock Nerf in the wild: Neural radiance fields for unconstrained photo
  collections.
\newblock {\em CVPR}, 2021.

\bibitem{mescheder2019occupancy}
Lars Mescheder, Michael Oechsle, Michael Niemeyer, Sebastian Nowozin, and
  Andreas Geiger.
\newblock Occupancy networks: Learning 3d reconstruction in function space.
\newblock In {\em Proceedings of the IEEE/CVF Conference on Computer Vision and
  Pattern Recognition}, pages 4460--4470, 2019.

\bibitem{meshry2019neural}
Moustafa Meshry, Dan~B Goldman, Sameh Khamis, Hugues Hoppe, Rohit Pandey, Noah
  Snavely, and Ricardo Martin-Brualla.
\newblock Neural rerendering in the wild.
\newblock In {\em Proceedings of the IEEE/CVF Conference on Computer Vision and
  Pattern Recognition}, pages 6878--6887, 2019.

\bibitem{mildenhall2020nerf}
Ben Mildenhall, Pratul~P Srinivasan, Matthew Tancik, Jonathan~T Barron, Ravi
  Ramamoorthi, and Ren Ng.
\newblock Nerf: Representing scenes as neural radiance fields for view
  synthesis.
\newblock In {\em European conference on computer vision}, pages 405--421.
  Springer, 2020.

\bibitem{Morad2021EmbodiedVN}
Steven~D. Morad, Roberto Mecca, Rudra P.~K. Poudel, Stephan Liwicki, and
  Roberto Cipolla.
\newblock Embodied visual navigation with automatic curriculum learning in real
  environments.
\newblock {\em IEEE Robotics and Automation Letters}, 6:683--690, 2021.

\bibitem{mueller2022instant}
Thomas M\"uller, Alex Evans, Christoph Schied, and Alexander Keller.
\newblock Instant neural graphics primitives with a multiresolution hash
  encoding.
\newblock {\em arXiv:2201.05989}, Jan. 2022.

\bibitem{niemeyer2021giraffe}
Michael Niemeyer and Andreas Geiger.
\newblock Giraffe: Representing scenes as compositional generative neural
  feature fields.
\newblock In {\em Proceedings of the IEEE/CVF Conference on Computer Vision and
  Pattern Recognition}, pages 11453--11464, 2021.

\bibitem{niemeyer2020differentiable}
Michael Niemeyer, Lars Mescheder, Michael Oechsle, and Andreas Geiger.
\newblock Differentiable volumetric rendering: Learning implicit 3d
  representations without 3d supervision.
\newblock In {\em Proceedings of the IEEE/CVF Conference on Computer Vision and
  Pattern Recognition}, pages 3504--3515, 2020.

\bibitem{Ost2021NeuralSG}
Julian Ost, Fahim Mannan, Nils Thuerey, Julian Knodt, and Felix Heide.
\newblock Neural scene graphs for dynamic scenes.
\newblock {\em 2021 IEEE/CVF Conference on Computer Vision and Pattern
  Recognition (CVPR)}, pages 2855--2864, 2021.

\bibitem{park2019deepsdf}
Jeong~Joon Park, Peter Florence, Julian Straub, Richard Newcombe, and Steven
  Lovegrove.
\newblock Deepsdf: Learning continuous signed distance functions for shape
  representation.
\newblock In {\em Proceedings of the IEEE/CVF Conference on Computer Vision and
  Pattern Recognition}, pages 165--174, 2019.

\bibitem{park2021nerfies}
Keunhong Park, Utkarsh Sinha, Jonathan~T Barron, Sofien Bouaziz, Dan~B Goldman,
  Steven~M Seitz, and Ricardo Martin-Brualla.
\newblock Nerfies: Deformable neural radiance fields.
\newblock In {\em Proceedings of the IEEE/CVF International Conference on
  Computer Vision}, pages 5865--5874, 2021.

\bibitem{peng2020convolutional}
Songyou Peng, Michael Niemeyer, Lars Mescheder, Marc Pollefeys, and Andreas
  Geiger.
\newblock Convolutional occupancy networks.
\newblock In {\em Computer Vision--ECCV 2020: 16th European Conference,
  Glasgow, UK, August 23--28, 2020, Proceedings, Part III 16}, pages 523--540.
  Springer, 2020.

\bibitem{pollefeys2008detailed}
Marc Pollefeys, David Nist{\'e}r, J-M Frahm, Amir Akbarzadeh, Philippos
  Mordohai, Brian Clipp, Chris Engels, David Gallup, S-J Kim, Paul Merrell,
  et~al.
\newblock Detailed real-time urban 3d reconstruction from video.
\newblock {\em IJCV}, 2008.

\bibitem{pumarola2021d}
Albert Pumarola, Enric Corona, Gerard Pons-Moll, and Francesc Moreno-Noguer.
\newblock D-nerf: Neural radiance fields for dynamic scenes.
\newblock In {\em Proceedings of the IEEE/CVF Conference on Computer Vision and
  Pattern Recognition}, pages 10318--10327, 2021.

\bibitem{ramon2021h3d}
Eduard Ramon, Gil Triginer, Janna Escur, Albert Pumarola, Jaime Garcia, Xavier
  Giro-i Nieto, and Francesc Moreno-Noguer.
\newblock H3d-net: Few-shot high-fidelity 3d head reconstruction.
\newblock In {\em Proceedings of the IEEE/CVF International Conference on
  Computer Vision}, pages 5620--5629, 2021.

\bibitem{reiser2021kilonerf}
Christian Reiser, Songyou Peng, Yiyi Liao, and Andreas Geiger.
\newblock {KiloNeRF}: Speeding up neural radiance fields with thousands of tiny
  {MLP}s.
\newblock {\em ICCV}, 2021.

\bibitem{rublee2011orb}
Ethan Rublee, Vincent Rabaud, Kurt Konolige, and Gary~R Bradski.
\newblock Orb: An efficient alternative to sift or surf.
\newblock In {\em ICCV}, 2011.

\bibitem{schonberger2016structure}
Johannes~L Schonberger and Jan-Michael Frahm.
\newblock Structure-from-motion revisited.
\newblock {\em CVPR}, 2016.

\bibitem{schwarz2020graf}
Katja Schwarz, Yiyi Liao, Michael Niemeyer, and Andreas Geiger.
\newblock Graf: Generative radiance fields for 3d-aware image synthesis.
\newblock {\em Advances in Neural Information Processing Systems},
  33:20154--20166, 2020.

\bibitem{shan2013turing}
Qi Shan, Riley Adams, Brian Curless, Yasutaka Furukawa, and Steven~M. Seitz.
\newblock The visual turing test for scene reconstruction.
\newblock {\em 3DV}, 2013.

\bibitem{sharma2022seeing}
Prafull Sharma, Ayush Tewari, Yilun Du, Sergey Zakharov, Rares Ambrus, Adrien
  Gaidon, William~T Freeman, Fredo Durand, Joshua~B Tenenbaum, and Vincent
  Sitzmann.
\newblock Seeing 3d objects in a single image via self-supervised
  static-dynamic disentanglement.
\newblock {\em arXiv preprint arXiv:2207.11232}, 2022.

\bibitem{snavely2006phototourism}
Noah Snavely, Steven~M. Seitz, and Richard Szeliski.
\newblock Photo tourism: Exploring photo collections in 3d.
\newblock {\em SIGGRAPH}, 2006.

\bibitem{sun2021direct}
Cheng Sun, Min Sun, and Hwann-Tzong Chen.
\newblock Direct voxel grid optimization: Super-fast convergence for radiance
  fields reconstruction.
\newblock {\em arXiv preprint arXiv:2111.11215}, 2021.

\bibitem{sun2022improved}
Cheng Sun, Min Sun, and Hwann-Tzong Chen.
\newblock Improved direct voxel grid optimization for radiance fields
  reconstruction.
\newblock {\em arXiv preprint arXiv:2206.05085}, 2022.

\bibitem{Takikawa2021NeuralGL}
Towaki Takikawa, Joey Litalien, K. Yin, Karsten Kreis, Charles~T. Loop, Derek
  Nowrouzezahrai, Alec Jacobson, Morgan McGuire, and Sanja Fidler.
\newblock Neural geometric level of detail: Real-time rendering with implicit
  3d shapes.
\newblock {\em 2021 IEEE/CVF Conference on Computer Vision and Pattern
  Recognition (CVPR)}, pages 11353--11362, 2021.

\bibitem{tancik2022block}
Matthew Tancik, Vincent Casser, Xinchen Yan, Sabeek Pradhan, Ben Mildenhall,
  Pratul~P Srinivasan, Jonathan~T Barron, and Henrik Kretzschmar.
\newblock Block-nerf: Scalable large scene neural view synthesis.
\newblock {\em arXiv preprint arXiv:2202.05263}, 2022.

\bibitem{tancik2021learned}
Matthew Tancik, Ben Mildenhall, Terrance Wang, Divi Schmidt, Pratul~P
  Srinivasan, Jonathan~T Barron, and Ren Ng.
\newblock Learned initializations for optimizing coordinate-based neural
  representations.
\newblock In {\em Proceedings of the IEEE/CVF Conference on Computer Vision and
  Pattern Recognition}, pages 2846--2855, 2021.

\bibitem{tancik2020fourfeat}
Matthew Tancik, Pratul~P. Srinivasan, Ben Mildenhall, Sara Fridovich-Keil,
  Nithin Raghavan, Utkarsh Singhal, Ravi Ramamoorthi, Jonathan~T. Barron, and
  Ren Ng.
\newblock Fourier features let networks learn high frequency functions in low
  dimensional domains.
\newblock {\em NeurIPS}, 2020.

\bibitem{tewari2021advances}
Ayush Tewari, Justus Thies, Ben Mildenhall, Pratul Srinivasan, Edgar Tretschk,
  Yifan Wang, Christoph Lassner, Vincent Sitzmann, Ricardo Martin-Brualla,
  Stephen Lombardi, et~al.
\newblock Advances in neural rendering.
\newblock {\em arXiv preprint arXiv:2111.05849}, 2021.

\bibitem{triggs1999bundle}
Bill Triggs, Philip~F McLauchlan, Richard~I Hartley, and Andrew~W Fitzgibbon.
\newblock Bundle adjustment—a modern synthesis.
\newblock {\em International workshop on vision algorithms}, 1999.

\bibitem{Truong2021BiDirectionalDA}
Joanne Truong, S. Chernova, and Dhruv Batra.
\newblock Bi-directional domain adaptation for sim2real transfer of embodied
  navigation agents.
\newblock {\em IEEE Robotics and Automation Letters}, 6:2634--2641, 2021.

\bibitem{turki2021mega}
Haithem Turki, Deva Ramanan, and Mahadev Satyanarayanan.
\newblock Mega-nerf: Scalable construction of large-scale nerfs for virtual
  fly-throughs.
\newblock {\em arXiv preprint arXiv:2112.10703}, 2021.

\bibitem{wang2021ibrnet}
Qianqian Wang, Zhicheng Wang, Kyle Genova, Pratul~P Srinivasan, Howard Zhou,
  Jonathan~T Barron, Ricardo Martin-Brualla, Noah Snavely, and Thomas
  Funkhouser.
\newblock Ibrnet: Learning multi-view image-based rendering.
\newblock In {\em Proceedings of the IEEE/CVF Conference on Computer Vision and
  Pattern Recognition}, pages 4690--4699, 2021.

\bibitem{xiangli2022bungeenerf}
Yuanbo Xiangli, Linning Xu, Xingang Pan, Nanxuan Zhao, Anyi Rao, Christian
  Theobalt, Bo Dai, and Dahua Lin.
\newblock Bungeenerf: Progressive neural radiance field for extreme multi-scale
  scene rendering.
\newblock In {\em European Conference on Computer Vision}, 2022.

\bibitem{xu2022point}
Qiangeng Xu, Zexiang Xu, Julien Philip, Sai Bi, Zhixin Shu, Kalyan Sunkavalli,
  and Ulrich Neumann.
\newblock Point-nerf: Point-based neural radiance fields.
\newblock {\em arXiv preprint arXiv:2201.08845}, 2022.

\bibitem{xu20213d}
Yinghao Xu, Sida Peng, Ceyuan Yang, Yujun Shen, and Bolei Zhou.
\newblock 3d-aware image synthesis via learning structural and textural
  representations.
\newblock {\em arXiv preprint arXiv:2112.10759}, 2021.

\bibitem{Yang2020SurfelGANSR}
Zhenpei Yang, Yuning Chai, Dragomir Anguelov, Yin Zhou, Pei Sun, D. Erhan, Sean
  Rafferty, and Henrik Kretzschmar.
\newblock Surfelgan: Synthesizing realistic sensor data for autonomous driving.
\newblock {\em 2020 IEEE/CVF Conference on Computer Vision and Pattern
  Recognition (CVPR)}, pages 11115--11124, 2020.

\bibitem{yu2021plenoxels}
Alex Yu, Sara Fridovich-Keil, Matthew Tancik, Qinhong Chen, Benjamin Recht, and
  Angjoo Kanazawa.
\newblock Plenoxels: Radiance fields without neural networks.
\newblock {\em arXiv preprint arXiv:2112.05131}, 2021.

\bibitem{yu2021plenoctrees}
Alex Yu, Ruilong Li, Matthew Tancik, Hao Li, Ren Ng, and Angjoo Kanazawa.
\newblock Plenoctrees for real-time rendering of neural radiance fields.
\newblock In {\em Proceedings of the IEEE/CVF International Conference on
  Computer Vision}, pages 5752--5761, 2021.

\bibitem{zhang2018unreasonable}
Richard Zhang, Phillip Isola, Alexei~A Efros, Eli Shechtman, and Oliver Wang.
\newblock The unreasonable effectiveness of deep features as a perceptual
  metric.
\newblock In {\em Proceedings of the IEEE conference on computer vision and
  pattern recognition}, pages 586--595, 2018.

\bibitem{zhang2022nerfusion}
Xiaoshuai Zhang, Sai Bi, Kalyan Sunkavalli, Hao Su, and Zexiang Xu.
\newblock Nerfusion: Fusing radiance fields for large-scale scene
  reconstruction.
\newblock {\em arXiv preprint arXiv:2203.11283}, 2022.

\bibitem{zhu2018very}
Siyu Zhu, Runze Zhang, Lei Zhou, Tianwei Shen, Tian Fang, Ping Tan, and Long
  Quan.
\newblock Very large-scale global {SFM} by distributed motion averaging.
\newblock {\em CVPR}, 2018.

\end{thebibliography}
}

\end{document}